\documentclass[sigconf]{acmart}
\AtBeginDocument{%
  }

\setcopyright{acmlicensed}

\copyrightyear{2026}
\acmYear{2026}
\setcopyright{cc}
\setcctype{by}
\acmConference[KDD 2026] {Proceedings of the 32nd ACM SIGKDD Conference on Knowledge Discovery and Data Mining V.2}{August 9--13, 2026}{Jeju Island, Republic of Korea.}
\acmBooktitle{Proceedings of the 32nd ACM SIGKDD Conference on Knowledge Discovery and Data Mining V.2 (KDD 2026), August 9--13, 2026, Jeju Island, Republic of Korea}
\acmISBN{979-8-4007-2259-2/2026/08}
\acmDOI{10.1145/3770855.3817631}
  
\usepackage{caption}
\usepackage{subcaption}
\usepackage{latexsym}
\usepackage{algorithm}
\usepackage{algorithmic}
\usepackage{graphicx}
\usepackage{booktabs}
\usepackage{algorithm}
\usepackage{algorithmic}
\usepackage{amsmath,bm}
\usepackage{multirow} 
\usepackage{multicol} 
\usepackage{xcolor}
\usepackage{colortbl}
\definecolor{Gray}{gray}{0.93}
\definecolor{Lavender}{rgb}{0.92,0.90,0.98}  % 柔和紫
\definecolor{SkyBlue}{rgb}{0.85,0.93,1.00}   % 或替代蓝色
\definecolor{LightCyan}{rgb}{0.88,1.00,1.00} % 或替代青色
\usepackage{mathrsfs}
\usepackage{tabularx}
\usepackage{makecell}
\usepackage{amsfonts}
\usepackage{bbm}
\usepackage{arydshln}
\usepackage{booktabs}
\usepackage{bbding}
\usepackage{pifont}
\begin{document}

% \title{The Name of the Title Is Hope}
\title[\methodname: Visual-Aware Adaptation of Large Language Models for Multilingual Web Image Translation]{\methodname: Visual-Aware Adaptation of Large Language Models for Multilingual Web Image Translation}

\author{Bo Li}
\affiliation{%
  \institution{Tsinghua University \\ Baidu Inc.}
  \state{Beijing}
  \country{China}
}
% \affiliation{%
%   \institution{Baidu Inc.}
%   \state{Beijing}
\authornote{Equal contribution}  

\author{Ronghao Chen}
\affiliation{%
  \institution{QuantaAlpha}
  \state{Beijing}
  \country{China}}
\authornotemark[1]

\author{Ningyuan Deng}
\affiliation{%
  \institution{The Hong Kong University of Science and Technology}
  \state{Hong Kong}
  \country{China}}

\author{Huacan Wang}
\affiliation{%
  \institution{QuantaAlpha}
  \state{Beijing}
  \country{China}
  }

\author{Shaolin Zhu}
\affiliation{%
  \institution{Tianjin University}
  \state{Tianjin}
  \country{China}
}
\authornote{Corresponding authors}

\author{Lijie Wen}
\affiliation{%
  \institution{Tsinghua University}
  \state{Beijing}
  \country{China}
}
\authornotemark[2]

\renewcommand{\shortauthors}{Bo Li et al.}

\newcommand{\methodname}{VaaWIT}
\newcommand{\llava}{LLaVA-OV}
\newcommand{\llama}{LLaMA3.1}
\newcommand{\mllama}{LLaMA3.2}
\newcommand{\qwenvl}{Qwen3-VL}
\newcommand{\qwen}{Qwen3}
\newcommand{\siglip}{SigLIP}
\newcommand{\msiglip}{mSigLIP}
\newcommand{\clip}{CLIP}
\newcommand{\blip}{BLIP-2}
\newcommand{\dino}{DINOv2}
\newcommand{\gptfour}{GPT4.1}
\newcommand{\gemini}{Gemini2.5}

%%数据集
\newcommand{\datasetmit}{MIT-10M}
\newcommand{\datasetecoit}{ECOIT}
\newcommand{\datasetopus}{OPUS-MIT-5M}

\begin{abstract}

Translating text embedded in Web images is crucial for improving content accessibility and cross-lingual information retrieval, particularly within social media and e-commerce domains.
Although Large Vision-Language Models (LVLMs) have advanced multimodal understanding, applying them to Web image translation remains challenging due to the visual representation gap: standard encoders often prioritize high-level semantics over the fine-grained visual details required for recognizing diverse character morphologies. 
To address this challenge, we propose \textbf{\methodname}, an end-to-end framework that adapts Large Language Models for multilingual Web image translation. 
The framework introduces two key technical contributions: (1) a Dual-Stream Attention Module (DSAM), which facilitates bidirectional interaction between multilingual semantic features and detailed visual representations, thereby synthesizing unified features robust to textual variations; and (2) a Visual-Aware Adapter (VAA), a parameter-efficient fine-tuning strategy that dynamically injects these fused visual cues into the frozen LLM backbone. 
This design enables the model to align the visual context with linguistic reasoning effectively while minimizing computational costs. 
Extensive experiments on eight tasks on three public benchmarks demonstrate that \methodname~significantly outperforms state-of-the-art (SOTA) open-source baselines and achieves competitive performance against proprietary models. 
These results validate the efficacy of integrating fine-grained visual perception into LLMs for complex Web content analysis.

\end{abstract}




\settopmatter{printacmref=true}


\maketitle

\section{Introduction}
\label{sec:intro}

Text embedded within Web images — ranging from e-commerce product descriptions and advertising posters to social media posts — serves as a primary carrier of information in the digital ecosystem. 
Unlike plain text, this visual text is characterized by diverse fonts, complex layouts, and significant background variations. 
Consequently, translating such content is critical for breaking language barriers in global information retrieval and content accessibility. 
However, this task presents unique challenges compared to standard neural machine translation (NMT), as it requires a system to simultaneously perform optical character recognition (OCR) and translation while preserving the semantic context provided by the visual scene \cite{mansimov2020towards, lan-etal-2024-translatotron}.
 
Existing approaches typically fall into two categories: cascaded systems and end-to-end specialized models. 
Cascaded systems, which sequentially apply OCR and NMT, suffer from error propagation; a recognition error in the OCR stage inevitably leads to translation failure \cite{yin2023multi}. 
While specialized end-to-end models \cite{zhu-etal-2023-peit, liang-etal-2024-document, niu-etal-2024-umtit} mitigate this issue by directly mapping image pixels to translated tokens, they often lack the scale and generalized world knowledge required to handle the linguistic diversity of the Web.

%%%%%%%%%%%%%%%%%%
\begin{figure*}[t]
    \centering
    \includegraphics[width=0.95\linewidth]{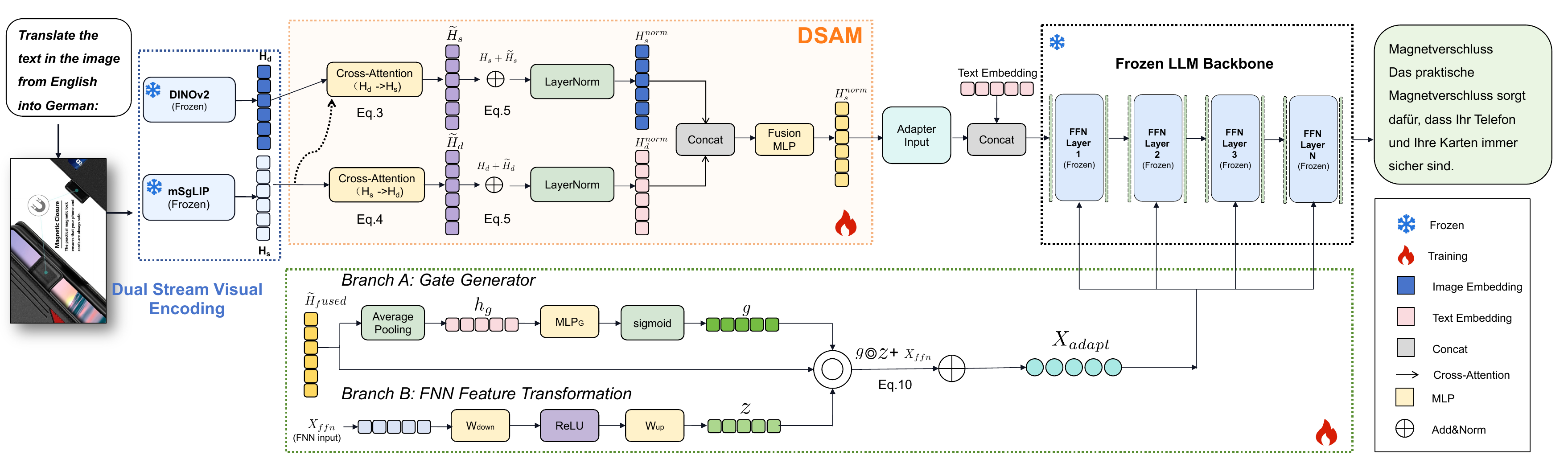}
    % \vspace{-24pt}
    \caption{Overview of \methodname. It addresses the complexity of Web image translation by decomposing the visual-linguistic alignment process into three integrated components: (1) \textit{Dual-Stream Visual Encoding}, (2) \textit{Visual Feature Fusion}, and (3) \textit{Visual-Aware LLM Adaptation}. We adopt a two-stage strategy to train the \methodname~framework: (1) Visual-Language Alignment and (2) Multi-Task Joint Learning.}
    % \vspace{-12pt}
    \label{fig:arch}
\end{figure*}
%%%%%%%%%%%%%%%%%%

Recently, Large Vision-Language Models (LVLMs) \cite{liu2024llavanext, lu2024deepseek, chen2024internvl, llama3modelcard, geminiteam2024gemini15unlockingmultimodal, li2023blip} have demonstrated remarkable capabilities in multimodal understanding. 
By aligning visual encoders with Large Language Models (LLMs), these architectures theoretically offer a unified solution for Web image translation.
Nevertheless, applying off-the-shelf LVLMs to this specific task exposes a critical Visual Representation Gap. Mainstream visual encoders (e.g., CLIP \cite{radford2021learning}) are optimized for image-level semantic alignment through contrastive learning. 
This pre-training paradigm encourages the encoder to capture high-level concepts (e.g., ``a red dress'') but often suppresses fine-grained visual details (e.g., the specific characters ``Sale 50\%" printed on the dress). 
This lack of morphological precision limits the ability of LLM to recognize and translate embedded text accurately~\cite{luo2024feast}. 
Furthermore, simply concatenating visual features with text prompts, a common fusion strategy, fails to establish a deep synergy between the visual details and the multilingual semantic context, resulting in hallucinations or omissions during translation~\cite{lin2023sphinx,jiang2023clip,shi2024eagle}.
Our ablation studies (Tables \ref{tab:ablation_study_detailed} and \ref{tab:fusion_strategy_comparison}) further validate this limitation empirically.

To address these limitations, we propose \methodname, an end-to-end framework designed to adapt LLMs for multilingual Web image translation. 
Unlike previous methods that rely on single-stream visual encoding, \methodname~effectively bridges the visual representation gap through two novel mechanisms. First, we introduce a Dual-Stream Attention Module (DSAM). 
This module processes visual inputs through two distinct pathways: a semantic stream (capturing global context) and a detail stream (capturing character morphology). 
A bidirectional cross-attention mechanism then fuses these streams, allowing semantic context to guide detail recognition and vice versa. 
Second, to integrate these fused representations into the LLM without incurring the high computational cost of full-parameter tuning, we design a Visual-Aware Adapter (VAA). 
This lightweight module dynamically modulates the LLM's internal representations based on visual cues, ensuring that the generation process is grounded in the visual evidence.
We conducted extensive experiments on 8 tasks with 3 public image translation tasks. 
The experimental results show that \methodname~substantially outperforms the SOTA open-source LVLMs such as \qwenvl~(32B) and \mllama~(90B), achieves performance comparable to \gptfour~and \gemini~Pro, and even surpasses them in several tasks.
 
Our contributions are summarized as follows.
\textbf{(I)} We identify the limitation of standard visual encoders in capturing text-centric visual details and propose \methodname, a framework that adapts LLMs for robust Web image translation through feature-level refinement. 
\textbf{(II)} We design the DSAM to synthesize fine-grained visual details with multilingual semantic context, and the VAA to enable parameter-efficient alignment between the vision module and the frozen LLM backbone.
\textbf{(III)} Extensive experiments on eight translation tasks in three benchmarks demonstrate that \methodname~significantly outperform open-source SOTA baselines and achieves performance competitive with proprietary commercial models, validating the effectiveness of our visual-aware adaptation strategy.

%%%%%%%%%%%%%%%%%%%%%%%%%%%%%%%%%%%%%%%%%%%%%%%%%%%%%%%%%%%%%%%%%%%%%%%%%%%%%%%
\section{Preliminaries}
\label{pre}

Let $\mathcal{D} = \{(\mathbf{X}_v^{(i)}, \mathbf{Y}^{(i)})\}_{i=1}^N$ denote a dataset comprising $N$ samples, where $\mathbf{X}_v \in \mathbb{R}^{H \times W \times 3}$ represents a raw Web image containing embedded text, and $\mathbf{Y} = \{y_1, y_2, \dots, y_L\}$ is the corresponding target translation sequence of length $L$. 
The objective is to learn a multimodal mapping function $\mathcal{M}$ that generates the target sequence $\mathbf{Y}$ conditioned on the visual input $\mathbf{X}_v$. 
We formulate this as an autoregressive generation problem, where the model maximizes the log-likelihood of the target tokens:

\begin{equation}
    \mathcal{L}(\theta) = \sum_{i=1}^N \sum_{j=1}^{L} \log P(y_j^{(i)} \mid y_{<j}^{(i)}, \mathbf{X}_v^{(i)}; \theta),
\end{equation}
where $\theta$ represents the trainable parameters and $y_{<j}$ denotes the tokens generated prior to the time step $j$. 
Unlike standard machine translation, which takes source text as input, our end-to-end setting requires the model to implicitly perform OCR and translation simultaneously based solely on pixel-level information.

To address the dual requirements of semantic understanding and character recognition in Web images, we leverage two distinct pre-trained visual backbones. 
First, a multilingual semantic encoder $\Phi_{sem}$ (e.g., SigLIP~\cite{zhai2023sigmoid}) is used to extract high-level semantic representations aligned with textual concepts, denoted $\mathbf{F}_{sem} = \Phi_{sem}(\mathbf{X}_v) \in \mathbb{R}^{N_v \times D_{sem}}$. 
While effective for global context, such encoders often lose high-frequency spatial information due to contrastive pre-training objectives. 
To compensate, we introduce a visual detail encoder $\Phi_{det}$ (e.g., DINOv2~\cite{oquab2023dinov2}), which is optimized by self-supervised learning to capture fine-grained morphological structures and layout details. 
This yields a detail-oriented feature set $\mathbf{F}_{det} = \Phi_{det}(\mathbf{X}_v) \in \mathbb{R}^{N_v \times D_{det}}$. 
Here, $N_v$ represents the number of visual patches, and $D_{sem}, D_{det}$ are the respective feature dimensions. 
These complementary feature streams serve as the input for our proposed Dual-Stream Attention Module.

%%%%%%%%%%%%%%%%%%%%%%%%%%%%%%%%%%%%%%%%%%%%%%%%%%%%%%%%%%%%%%%%%%%%%%%%%%%%%%%%
\section{Methodology}
 
This section details the architecture and optimization strategy of \methodname~framework. 
As illustrated in Figure~\ref{fig:arch}, \methodname~is designed to bridge the gap between fine-grained visual perception and multilingual semantic reasoning through a unified end-to-end pipeline.

\subsection{Framework Overview}
\label{sec:overview}

The proposed framework addresses the complexity of Web image translation by decomposing the visual-linguistic alignment process into three integrated components: (1) \textit{Dual-Stream Visual Encoding}, (2) \textit{Visual Feature Fusion}, and (3) \textit{Visual-Aware LLM Adaptation}.

\noindent \textbf{Dual-Stream Visual Encoding.} Given an input image $\mathbf{X}_v$, the system first extracts complementary visual representations to capture both high-level semantics and low-level morphological details. 
As defined in Section \ref{pre}, we employ visual encoders consisting of a multilingual semantic encoder ($\Phi_{sem}$) and a visual detail encoder ($\Phi_{det}$). 
These encoders operate in parallel to produce the semantic feature sequence $\mathbf{F}_{sem}$ and the detail feature sequence $\mathbf{F}_{det}$, respectively.

\noindent \textbf{Dual-Stream Attention Module (DSAM).} To synthesize these heterogeneous features, the DSAM facilitates bidirectional interaction between $\mathbf{F}_{sem}$ and $\mathbf{F}_{det}$. 
Through a symmetric cross-attention mechanism, semantic context is used to filter and refine morphological details, while fine-grained visual cues enhance semantic clarity. 
This process yields a unified visual representation, denoted as $\mathbf{H}_{fused}$, which is robust to visual noise and stylistic variations inherent in Web images.

\noindent \textbf{Visual-Aware Adapter (VAA).} To effectively leverage $\mathbf{H}_{fused}$ for translation without compromising the linguistic generalization of the LLM, we introduce the Visual-Aware Adapter network. 
Unlike static prefix tuning, VAA injects visual information into the intermediate layers of the frozen LLM backbone ($\Psi_{LLM}$) via a dynamic gating mechanism. 
This allows the model to adaptively modulate its hidden states conditioned on visual evidence during the auto-regressive generation of the target translation $\mathbf{Y}$.
  
\subsection{Dual-Stream Attention Module (DSAM)}
\label{sec:dsam}

The DSAM serves as the core fusion engine, designed to bridge the modality gap between high-level semantics and fine-grained visual details. 
As illustrated in Figure~\ref{fig:arch}, DSAM takes the outputs from the semantic and detail encoders as input and synthesizes a unified visual representation.
First, given the raw feature sequences $\mathbf{F}_{sem} \in \mathbb{R}^{N_v \times D_{sem}}$ and $\mathbf{F}_{det} \in \mathbb{R}^{N_v \times D_{det}}$ extracted by the visual encoders, we project them into a shared latent space with dimension $D$. 
This is achieved via linear transformations:

\begin{equation}
    \mathbf{H}_{s} = \mathbf{F}_{sem} \mathbf{W}_{s}, \quad \mathbf{H}_{d} = \mathbf{F}_{det} \mathbf{W}_{d},
\end{equation}
where $\mathbf{W}_{s} \in \mathbb{R}^{D_{sem} \times D}$ and $\mathbf{W}_{d} \in \mathbb{R}^{D_{det} \times D}$ are learnable projection matrices. $\mathbf{H}_{s}$ and $\mathbf{H}_{d}$ represent the projected semantic and detail feature sequences, respectively.
A naive concatenation of $\mathbf{H}_{s}$ and $\mathbf{H}_{d}$ is insufficient to capture the intricate dependencies between textual semantics and visual morphology. 
To address this, we employ \textit{Semantic-Guided Detail Refinement (SGDR)} and \textit{Detail-Informed Semantic Refinement (DISR)} that allow each stream to query information from the other. 
Specifically, the SGDR uses semantic features as the query to retrieve relevant morphological details:

\begin{equation}
    \tilde{\mathbf{H}}_{d} = \text{MHA}(\mathbf{Q}=\mathbf{H}_{s}, \mathbf{K}=\mathbf{H}_{d}, \mathbf{V}=\mathbf{H}_{d}),
\end{equation}
where MHA denotes Multi-Head Attention. 
Conversely, the DISR enhances semantic features with precise visual cues:

\begin{equation}
    \tilde{\mathbf{H}}_{s} = \text{MHA}(\mathbf{Q}=\mathbf{H}_{d}, \mathbf{K}=\mathbf{H}_{s}, \mathbf{V}=\mathbf{H}_{s}).
\end{equation}
Here, $\tilde{\mathbf{H}}_{d}$ represents detail features reorganized by semantic context (e.g., focusing on text regions identified by semantics), while $\tilde{\mathbf{H}}_{s}$ denotes semantic features enriched with fine-grained visual evidence.

Following the attention layers, we apply residual connections and Layer Normalization (LN) to stabilize the gradients:
\begin{equation}
    \hat{\mathbf{H}}_{s} = \text{LN}(\mathbf{H}_{s} + \tilde{\mathbf{H}}_{s}), \quad \hat{\mathbf{H}}_{d} = \text{LN}(\mathbf{H}_{d} + \tilde{\mathbf{H}}_{d}).
\end{equation}
Finally, the refined features from both streams are concatenated and fused through a Multi-Layer Perceptron (MLP) to produce the final visual representation sequence:
\begin{equation}
    \mathbf{H}_{fused} = \text{MLP}_{fusion}([\hat{\mathbf{H}}_{s}; \hat{\mathbf{H}}_{d}]) \in \mathbb{R}^{N_v \times D_{LLM}},
\end{equation}
where $[\cdot;\cdot]$ denotes concatenation along the feature dimension, and $D_{LLM}$ aligns with the hidden dimension of the LLM backbone. 
This fused representation $\mathbf{H}_{fused}$ effectively encapsulates both the linguistic context required for translation and the visual details necessary for character recognition.
 
\subsection{Visual-Aware Adapter (VAA)}
\label{sec:vaa}

Standard adaptation methods often treat visual inputs as static prefixes, which may not effectively modulate the generative process of LLMs when dealing with varying visual complexities. 
To address this, we propose the VAA, a lightweight module injected into the transformer layers of the frozen LLM backbone. 
VAA dynamically regulates the infusion of visual information via a content-dependent gating mechanism.
Since the fused visual sequence $\mathbf{H}_{fused} \in \mathbb{R}^{N_v \times D_{LLM}}$ contains dense patch-level information, directly injecting it into every layer incurs significant computational overhead. 
Instead, we first aggregate the sequence into a global visual context vector $\mathbf{h}_{g}$ via average pooling:

\begin{equation}
    \mathbf{h}_{g} = \frac{1}{N_v} \sum_{i=1}^{N_v} \mathbf{H}_{fused}^{(i)},
\end{equation}
where $\mathbf{H}_{fused}^{(i)}$ denotes the feature vector of the $i$-th visual patch. 
This global vector encapsulates the overall semantic and stylistic essence of the input image.
Within each transformer layer $l$, the VAA operates on the output of the Feed-Forward Network (FFN), denoted as $\mathbf{x}^{(l)}_{ffn}$. 
To dynamically control the influence of visual context, we employ a gating network $\mathcal{G}$ that computes a soft gate vector $\mathbf{g} \in (0, 1)^{D_{LLM}}$ conditioned on the global visual context:

\begin{equation}
    \mathbf{g} = \sigma(\text{MLP}_{\mathcal{G}}(\mathbf{h}_{g})),
\end{equation}
where $\sigma(\cdot)$ is the element-wise sigmoid function. 

Concurrently, a bottleneck adapter transforms the layer activation $\mathbf{x}^{(l)}_{ffn}$. 
Following the standard bottleneck design~\cite{houlsby2019parameter}, the adapter consists of a down-projection $\mathbf{W}_{down} \in \mathbb{R}^{D_{LLM} \times r}$ and an up-projection $\mathbf{W}_{up} \in \mathbb{R}^{r \times D_{LLM}}$, where $r \ll D_{LLM}$ is the bottleneck dimension:

\begin{equation}
    \mathbf{z}^{(l)} = \mathbf{W}_{up} \cdot \text{ReLU}(\mathbf{W}_{down} \cdot \mathbf{x}^{(l)}_{ffn}).
\end{equation}
The gated visual adaptation is then applied via element-wise multiplication:
\begin{equation}
    \mathbf{x}^{(l)}_{adapt} = \mathbf{x}^{(l)}_{ffn} + \mathbf{g} \odot \mathbf{z}^{(l)}.
\end{equation}
Here, the residual connection ensures that the pre-trained linguistic knowledge is preserved, while the gate $\mathbf{g}$ allows the model to selectively enhance or suppress visual adaptation based on the confidence of the visual signal.

The final output of the transformer layer $l$ is obtained by adding the gated adapter output to the residual stream. 
This design enables the LLM to perform visual-aware reasoning while maintaining parameter efficiency, as only the lightweight adapter weights and the gating network are updated during training.
 
%%%%%%%%%%%%%%%%%%%%%%%%%%%%%%%%%%%%%%%%%%%%%%%%%%%%%%
\begin{table*}[t]
\centering
\caption{
\textbf{Comparison with Cascaded pipelines, SOTA LVLMs (Zero-Shot), and various tuning strategies (based on \qwenvl)}.
Best results in each column are \textbf{bolded}.
The last two rows (highlighted in Gray) represent \methodname~with different LLM backbones (\qwen~and \llama).
Subscripts denote standard deviations: values are computed over three random seeds. \textsuperscript{$\dagger$}Commercial model evaluations were conducted in October 2025.
}
\resizebox{\textwidth}{!}{
\begin{tabular}{lcccccccccccccccc}
  \toprule
  & \multicolumn{2}{c}{ \bf ZH-EN} & \multicolumn{2}{c}{ \bf EN-IT} & \multicolumn{2}{c}{ \bf EN-JA} & \multicolumn{2}{c}{ \bf IT-EN} & \multicolumn{2}{c}{ \bf JA-EN} & \multicolumn{2}{c}{ \bf HI-EN} & \multicolumn{2}{c}{ \bf KO-EN} & \multicolumn{2}{c}{ \bf TH-EN} \\
   \cmidrule(lr){2-3} \cmidrule(lr){4-5} \cmidrule(lr){6-7} \cmidrule(lr){8-9} \cmidrule(lr){10-11} \cmidrule(lr){12-13} \cmidrule(lr){14-15} \cmidrule(lr){16-17}
   & {\footnotesize BLEU} & {\footnotesize COMET} & {\footnotesize BLEU} & {\footnotesize COMET} & {\footnotesize BLEU} & {\footnotesize COMET} & {\footnotesize BLEU} & {\footnotesize COMET} & {\footnotesize BLEU} & {\footnotesize COMET} & {\footnotesize BLEU} & {\footnotesize COMET} & {\footnotesize BLEU} & {\footnotesize COMET} & {\footnotesize BLEU} & {\footnotesize COMET}  \\
  \midrule
  \multicolumn{17}{c}{\textit{Cascaded Models}} \\ 
  \midrule
EasyOCR\_Google API  & 9.9$_{\pm.8}$ & 63.5$_{\pm1.2}$ & 25.0$_{\pm1.1}$ & 69.2$_{\pm1.0}$ & 8.6$_{\pm.7}$ & 60.8$_{\pm1.1}$ & 5.6$_{\pm.6}$ & 50.7$_{\pm1.3}$ & 9.1$_{\pm.7}$ & 56.8$_{\pm1.2}$ & 21.4$_{\pm1.0}$ & 65.3$_{\pm1.1}$ & 33.9$_{\pm1.2}$ & 86.1$_{\pm.8}$ & 20.3$_{\pm1.0}$ & 69.8$_{\pm1.0}$ \\
PP-OCR\_Microsoft API & 9.6$_{\pm.8}$ & 63.8$_{\pm1.2}$ & 27.5$_{\pm1.1}$ & 71.9$_{\pm1.0}$ & 11.2$_{\pm.8}$ & 62.4$_{\pm1.1}$ & 4.7$_{\pm.5}$ & 47.3$_{\pm1.4}$ & 5.2$_{\pm.6}$ & 46.6$_{\pm1.3}$ & 15.4$_{\pm.9}$ & 67.4$_{\pm1.1}$ & 27.1$_{\pm1.1}$ & 81.3$_{\pm.9}$ & 23.5$_{\pm1.0}$ & 76.5$_{\pm.9}$ \\
  \midrule
  \multicolumn{17}{c}{\textit{SOTA LVLMs Zero-Shot}} \\ 
  \midrule
\qwenvl~(8B) & 32.3$_{\pm1.3}$ & 81.3$_{\pm.9}$ & 22.7$_{\pm1.1}$ & 68.8$_{\pm1.0}$ & 11.4$_{\pm.8}$ & 55.6$_{\pm1.2}$ & 32.8$_{\pm1.2}$ & 74.7$_{\pm1.0}$ & 26.1$_{\pm1.1}$ & 67.2$_{\pm1.0}$ & 10.1$_{\pm.8}$ & 63.1$_{\pm1.1}$ & 27.9$_{\pm1.1}$ & 69.6$_{\pm1.1}$ & 12.2$_{\pm.8}$ & 64.8$_{\pm1.1}$ \\
\qwenvl~(32B)  & 39.2$_{\pm1.4}$ & 84.4$_{\pm.8}$ & 34.1$_{\pm1.2}$ & 76.8$_{\pm.9}$ & 15.5$_{\pm.9}$ & 55.9$_{\pm1.2}$ & 48.7$_{\pm1.3}$ & 80.5$_{\pm.9}$ & 26.9$_{\pm1.1}$ & 65.8$_{\pm1.1}$ & 10.5$_{\pm.8}$ & 63.3$_{\pm1.1}$ & 33.9$_{\pm1.2}$ & 77.9$_{\pm1.0}$ & 14.3$_{\pm.9}$ & 69.9$_{\pm1.0}$ \\
\mllama~(11B) & 2.8$_{\pm.4}$ & 45.7$_{\pm1.4}$ & 2.9$_{\pm.4}$ & 48.7$_{\pm1.3}$ & 1.6$_{\pm.3}$ & 50.7$_{\pm1.3}$ & 10.1$_{\pm.8}$ & 61.4$_{\pm1.2}$ & 3.1$_{\pm.4}$ & 43.1$_{\pm1.4}$ & 2.1$_{\pm.4}$ & 45.8$_{\pm1.4}$ & 5.0$_{\pm.6}$ & 52.6$_{\pm1.3}$ & 3.1$_{\pm.4}$ & 47.9$_{\pm1.4}$ \\
\mllama~(90B) & 7.9$_{\pm.7}$ & 48.9$_{\pm1.4}$ & 9.2$_{\pm.7}$ & 51.6$_{\pm1.3}$ & 3.6$_{\pm.5}$ & 54.8$_{\pm1.2}$ & 12.4$_{\pm.8}$ & 50.3$_{\pm1.3}$ & 14.8$_{\pm.9}$ & 49.3$_{\pm1.3}$ & 2.8$_{\pm.4}$ & 49.2$_{\pm1.3}$ & 6.9$_{\pm.7}$ & 65.8$_{\pm1.1}$ & 5.5$_{\pm.6}$ & 48.0$_{\pm1.4}$ \\
\llava~(7B) & 1.7$_{\pm.3}$ & 42.6$_{\pm1.5}$ & 6.2$_{\pm.6}$ & 54.1$_{\pm1.2}$ & 3.4$_{\pm.5}$ & 49.7$_{\pm1.3}$ & 12.2$_{\pm.8}$ & 62.9$_{\pm1.2}$ & 11.2$_{\pm.8}$ & 47.5$_{\pm1.3}$ & 1.6$_{\pm.3}$ & 44.2$_{\pm1.4}$ & 3.0$_{\pm.4}$ & 46.1$_{\pm1.3}$ & 2.4$_{\pm.4}$ & 44.2$_{\pm1.5}$ \\
\rowcolor{Lavender} \gptfour\textsuperscript{$\dagger$} & 46.1$_{\pm1.4}$ & 94.6$_{\pm.4}$ & 30.0$_{\pm1.2}$ & 75.2$_{\pm.9}$ & 23.7$_{\pm1.1}$ & 81.6$_{\pm.8}$ & 48.6$_{\pm1.3}$ & 89.4$_{\pm.6}$ & 25.8$_{\pm1.1}$ & 68.2$_{\pm1.0}$ & 17.3$_{\pm.9}$ & 69.9$_{\pm1.0}$ & \textbf{43.9}$_{\pm1.3}$ & \textbf{94.6}$_{\pm.4}$ & \textbf{43.2}$_{\pm1.2}$ & \textbf{96.2}$_{\pm.3}$ \\
\rowcolor{Lavender} \gemini~Pro\textsuperscript{$\dagger$} & 40.1$_{\pm1.4}$ & 90.1$_{\pm.6}$ & 31.5$_{\pm1.2}$ & 70.3$_{\pm1.0}$ & 22.2$_{\pm1.0}$ & 75.5$_{\pm.9}$ & 40.8$_{\pm1.3}$ & 76.9$_{\pm.9}$ & 20.9$_{\pm1.0}$ & 66.1$_{\pm1.1}$ & 17.5$_{\pm.9}$ & 70.9$_{\pm1.0}$ & 42.6$_{\pm1.3}$ & 87.1$_{\pm.7}$ & 42.1$_{\pm1.2}$ & 91.1$_{\pm.5}$ \\
  \midrule
  \multicolumn{17}{c}{\textit{\qwenvl~(8B) Tuning Strategies}} \\ 
  \midrule
Chain-of-Thought & 32.7$_{\pm1.3}$ & 81.7$_{\pm.9}$ & 32.2$_{\pm1.2}$ & 70.3$_{\pm1.0}$ & 13.5$_{\pm.9}$ & 67.7$_{\pm1.0}$ & 18.6$_{\pm1.0}$ & 65.3$_{\pm1.1}$ & 26.2$_{\pm1.1}$ & 67.0$_{\pm1.0}$ & 14.7$_{\pm.9}$ & 69.0$_{\pm1.0}$ & 27.5$_{\pm1.1}$ & 77.1$_{\pm.9}$ & 18.6$_{\pm1.0}$ & 71.0$_{\pm1.0}$ \\
LoRA & 35.0$_{\pm.5}$ & 83.6$_{\pm.6}$ & 32.9$_{\pm.4}$ & 67.7$_{\pm.5}$ & 9.0$_{\pm.4}$ & 58.8$_{\pm.6}$ & 27.5$_{\pm.5}$ & 75.4$_{\pm.5}$ & 33.1$_{\pm.5}$ & 73.9$_{\pm.5}$ & 11.1$_{\pm.4}$ & 65.2$_{\pm.6}$ & 19.2$_{\pm.5}$ & 75.2$_{\pm.5}$ & 10.5$_{\pm.4}$ & 62.4$_{\pm.6}$ \\
Full Fine-Tuning & 61.8$_{\pm.6}$ & 89.6$_{\pm.5}$ & 36.9$_{\pm.5}$ & 72.4$_{\pm.5}$ & 15.3$_{\pm.5}$ & 51.0$_{\pm.6}$ & 58.4$_{\pm.5}$ & 85.3$_{\pm.5}$ & 45.0$_{\pm.5}$ & 84.3$_{\pm.5}$ & 34.7$_{\pm.5}$ & 73.0$_{\pm.5}$ & 31.4$_{\pm.5}$ & 83.1$_{\pm.5}$ & 36.8$_{\pm.5}$ & 84.4$_{\pm.5}$ \\
  \midrule
  \multicolumn{17}{c}{\textit{Ours \methodname}} \\ 
  \midrule
\rowcolor{Gray} \small \textbf{\methodname~on \qwen~(8B)} & \textbf{65.9}$_{\pm.4}$ & \textbf{94.8}$_{\pm.5}$ & \textbf{38.1}$_{\pm.3}$ & \textbf{85.9}$_{\pm.4}$ & \textbf{26.4}$_{\pm.5}$ & \textbf{83.9}$_{\pm.6}$ & 66.0$_{\pm.3}$ & 88.2$_{\pm.4}$ & 47.2$_{\pm.4}$ & 86.6$_{\pm.5}$ & \textbf{41.3}$_{\pm.5}$ & \textbf{85.0}$_{\pm.4}$ & 35.2$_{\pm.3}$ & 85.6$_{\pm.5}$ & 39.8$_{\pm.4}$ & 87.5$_{\pm.5}$ \\
\rowcolor{Gray} \small \textbf{\methodname~on \llama~(8B)} & 59.3$_{\pm.5}$ & 88.7$_{\pm.6}$ & 35.2$_{\pm.4}$ & 81.6$_{\pm.5}$ & 26.2$_{\pm.4}$ & 83.1$_{\pm.5}$ & \textbf{78.1}$_{\pm.4}$ & \textbf{94.8}$_{\pm.3}$ & \textbf{55.3}$_{\pm.5}$ & \textbf{94.6}$_{\pm.4}$ & 35.7$_{\pm.4}$ & 83.9$_{\pm.5}$ & 36.2$_{\pm.3}$ & 88.7$_{\pm.4}$ & 33.5$_{\pm.5}$ & 76.4$_{\pm.6}$ \\
  \bottomrule
\end{tabular}
}
\label{tab:main_results}
\end{table*}
%%%%%%%%%%%%%%%%%%%%%%%%%%%%%%%%%%%%%%%%%%%%%%%%%%%%%%
\subsection{Training}
\label{sec:training}

Training of \methodname~follows a two-stage paradigm designed to progressively align visual perception with linguistic generation. 
Throughout both stages, the parameters of the visual encoders and the LLM backbone remain frozen, while only the DSAM and VAA modules are updated.

\textbf{Stage 1: Visual-Language Alignment.} The primary goal of this stage is to initialize the newly introduced modules by aligning the fused visual representation $\mathbf{H}_{fused}$ with the LLM's semantic space. 
We treat this as a standard image captioning task, where the model learns to reconstruct the text contained in the image. 
Let $\mathbf{Y}$ denote the ground-truth text sequence. The alignment loss is defined as the negative log-likelihood:

\begin{equation}
    \mathcal{L}_{align} = -\sum_{j=1}^{L} \log P(y_j \mid y_{<j}, \mathbf{H}_{fused}; \theta),
\end{equation}
where $\theta$ denotes the trainable parameters of DSAM and VAA. 
This stage ensures that the visual features provide a reliable starting point for the subsequent translation task.

\textbf{Stage 2: Multi-Task Joint Learning.} To robustly handle the complexities of Web image translation, we fine-tune the model using a multi-task learning objective. 
This stage integrates three complementary tasks:
% \begin{itemize}
%     \item 
\textit{Image-Text Matching (ITM):} To enforce global semantic consistency, the model predicts whether a given text sequence matches the visual content.
This is formulated as a binary classification task conditioned on $\mathbf{H}_{fused}$.
    % \item 
\textit{Text Translation Learning (TTL):} To maintain the LLM's inherent machine translation capabilities, we include a pure text-to-text translation task. 
The model generates the target translation $\mathbf{Y}$ given the source text $\mathbf{T}^s$, optimizing $\mathcal{L}_{TTL} = -\log P(\mathbf{Y} \mid \mathbf{T}^s)$.
    % \item 
\textit{Image Translation Learning (ITL):} This is the core task. The model generates the target translation $\mathbf{Y}$ conditioned on both the visual representation $\mathbf{H}_{fused}$ and the source text $\mathbf{T}^s$. 
The objective is $\mathcal{L}_{ITL} = -\log P(\mathbf{Y} \mid \mathbf{T}^s, \mathbf{H}_{fused})$.
% \end{itemize}

The final objective function is a weighted sum of these components:
\begin{equation}
    \mathcal{L}_{total} = \lambda_{ITM} \mathcal{L}_{ITM} + \lambda_{TTL} \mathcal{L}_{TTL} + \lambda_{ITL} \mathcal{L}_{ITL},
\end{equation}
where $\lambda_{ITM}, \lambda_{TTL}, \lambda_{ITL}$ are hyperparameters balancing the contribution of semantic alignment, linguistic fluency, and multimodal translation, respectively. 
Empirically, we set $\lambda_{ITL} > \lambda_{ITM} > \lambda_{TTL}$ to prioritize the end-to-end translation performance.
Hyperparameters and optimization details are summarized in Table~\ref{tab:hyperparameters}.

\section{Experiments}

\subsection{Experimental Setup}

\noindent \textbf{Datasets.}
To comprehensively evaluate our approach, we conducted experiments on 3 public Web image translation datasets covering 8 tasks.
\textbf{\datasetmit~\cite{li-etal-2025-mit}} is a large-scale dataset of multilingual Web images collected from real-world websites. 
We selected four tasks (EN-IT, IT-EN, EN-JA, and JA-EN).
\textbf{\datasetecoit~\cite{zhu-etal-2023-peit}} contains product images from Chinese e-commerce websites (ZH-EN).
\textbf{\datasetopus}~\cite{li2026mnaft} is a multilingual synthetic dataset simulating social media meme-style images. 
We selected three tasks (HI-EN, KO-EN, and TH-EN).
The tasks selected from each dataset aim to cover both High-resource languages (English (EN), Italian (IT)) and Lower-resource languages (Chinese (ZH), Japanese (JA), Korean (KO), Thai (TH), Hindi (HI)). 

We use BLEU (SacreBLEU) \cite{papineni2002bleu}, which is widely used in the field of machine translation, and COMET \cite{rei-etal-2020-comet} \footnote{https://huggingface.co/Unbabel/wmt22-comet-da}, an automatic evaluation metric based on neural networks, to evaluate the accuracy of our method. 
We aim to provide a comprehensive assessment of Web image translation quality in terms of both surface similarity and semantic fidelity.

\noindent \textbf{Baselines.}
We compared \methodname~against cascaded systems and SOTA E2E models.
The cascaded model first applies EasyOCR \footnote{\url{https://github.com/JaidedAI/EasyOCR}} or PP-OCR~\cite{li2022pp} extracts text from images and then translates the extracted text using the Google and Microsoft Translate APIs.
This choice of established components makes our baseline representative of typical cascaded methods and facilitates reproducibility.
And we compared \methodname~with SOTA LVLMs (Zero-Shot):
\qwenvl~(8B,32B) \cite{bai2025qwen3vltechnicalreport}, \llava~(7B) \cite{li2024llava}, \mllama~(11B,70B) \cite{grattafiori2024llama}, \gptfour~\cite{achiam2023gpt}, \gemini~Pro \cite{DeepMind_GeminiPro} and various tuning strategies of LVLMs for Web image translation: Chain-of-Thought (CoT) \cite{NEURIPS2022_9d560961}, LoRA \cite{hu2022lora}, Full Fine-tuning.
For the E2E IT model, we compared \methodname~with the latest image translation methods ItNet \cite{jain2021image}, E2ETIT \cite{Ma2022ImprovingET}, PEIT \cite{zhu-etal-2023-peit}, Translatotron-V \cite{lan-etal-2024-translatotron}, AnyTrans \cite{qian-etal-2024-anytrans} and DIMTDA \cite{liang-etal-2024-document}.
The detailed experimental settings and the list of baseline methods are provided in Appendix \ref{sec:implementation_details}.

%%%%%%%%%%%%%%%%%%%%%%%%%%%%%%%%%%%%%%%%%%%%%%%%%%%%%%%%%%%%%%
\begin{table}[t]
\centering
\caption{
Comparison of trainable parameters.
}
\label{tab:parameter_efficiency}
\resizebox{0.85\linewidth}{!}{
\begin{tabular}{lr}
\toprule
\textbf{Method} & \textbf{Trainable Parameters}  \\
\midrule
Full Fine-Tuning & 8B  \\
LoRA ($r$=8) & 30M \\
\textbf{\methodname~} (DSAM + VAA, frozen LLM) & \textbf{50M} \\
\bottomrule
\end{tabular}
}
\vspace{-2mm}
\end{table}
%%%%%%%%%%%%%%%%%%%%%%%%%%%%%%%%%%%%%%%%%%%%%%%%%%%%%%%%%%%%%%

\subsection{Main Results}

We conducted a comprehensive evaluation of \methodname~in 8 tasks (ZH-EN, EN-IT, EN-JA, IT-EN, JA-EN, HI-EN, KO-EN, TH-EN), comparing it with a wide range of standard methods, including traditional cascaded pipelines, SOTA LVLMs (Zero-Shot) and various fine-tuning adaptation strategies.
We implemented and tested \methodname~on two LLM backbones: \qwen~(8B) and \llama~(8B), evaluating its consistency and transferability between different LLM architectures.
Detailed results are presented in Table~\ref{tab:main_results}.

Compared to traditional cascade models, \methodname~achieves significant improvements in all language pairs.
For example, on the ZH-EN task, \methodname~surpasses the combinations of EasyOCR and Google Translate API and PP-OCR and Microsoft Translator API by more than 50 BLEU points, demonstrating the advantage of its end-to-end design in eliminating error propagation and capturing multimodal contextual information.
More compellingly, \methodname~substantially outperforms SOTA LVLMs such as \mllama~(90B) (Zero-Shot), and we also compared our model with leading commercial closed-source systems.
Across most tasks, \methodname~achieves performance comparable to \gptfour~and \gemini~Pro, and even surpasses them on several tasks.
These results show that even highly capable general purpose LVLMs still face limitations when dealing with the complex visual–semantic alignment challenges of Web image translation, while \methodname, through its design, achieves a superior balance between semantic understanding and fine-grained visual features, highlighting both the difficulty of the task and the effectiveness of our approach.

We further compared \methodname~with several adaptation strategies (based on \qwenvl), including Chain-of-Thought (CoT), LoRA, and Full FT.
The results show that simple prompting or lightweight tuning yields only limited improvement, while full fine-tuning achieves stronger results at a much higher computational cost.
In contrast, \methodname~trains only the lightweight DSAM and VAA (around 50M trainable parameters), but still surpasses the fully fine-tuned models on most tasks.
For example, on IT-EN, \methodname~achieves 66.0 BLEU / 88.2 COMET, improving over Full FT by 7.6 BLEU and 2.9 COMET.

During training, all parameters of the visual encoders and LLM backbone are frozen, and only the lightweight DSAM and VAA modules are updated.
Both modules are designed to be parameter efficient, making \methodname's training cost far lower than training or fully fine-tuning an LVLM from scratch.
As summarized in Table~\ref{tab:parameter_efficiency}, \methodname~has approximately 50M trainable parameters, on the same order of magnitude as LoRA (30M), but much fewer than Full FT (8B).
And the two-stage training stage (one epoch per stage) requires only about 18 hours.
Under these highly efficient conditions, \methodname~still outperforms large models such as \gptfour~and \gemini~Pro in multiple tasks, clearly demonstrating its superior performance–efficiency trade-off.

Moreover, \methodname~achieves consistently strong results on both both \qwen~and \llama~LLM backbones, further confirming the robust generalization and scalability of the framework across different LLM architectures.
In summary, \methodname~exhibits consistent and powerful performance in multilingual Web image translation tasks.
With excellent parameter efficiency and an acceptable training cost, it proves to be an effective and scalable solution for real-world multilingual Web image translation.

\subsection{Comparison with Image Translation Models}

%%%%%%%%%%%%%%%%%%%%%%%%%%%%%%%%%%
\begin{table}[t]
\centering
% \small
\caption{Comparison with the SOTA image translation models (LLM Backbone: \qwen-8B)  in ZH-EN and EN-IT tasks. 
% We highlight the best numbers in \textbf{bold}.
}
\label{tab:translation_sota_results}
\resizebox{0.95\linewidth}{!}{
\begin{tabular}{lrcccc}
\toprule
& &\multicolumn{2}{c}{\bf ZH-EN} & \multicolumn{2}{c}{\bf EN-IT} \\
 \cmidrule(r){3-4} \cmidrule(l){5-6}
& \bf \#param  & BLEU & COMET & BLEU & COMET \\
\midrule
% \rowcolor{Gray}
\bf \methodname~(Ours) &  \bf 50M & \bf 65.9 & \bf 94.8 &  \bf 38.1 & \bf 85.9 \\
\midrule
ItNet \cite{jain2021image} & 60.6M & 39.3 & 71.1 & 25.1 & 58.9 \\
PEIT \cite{zhu-etal-2023-peit}&  71.6M & 47.2 & 79.2 & 30.9 & 72.0 \\
Translatotron-V \cite{lan-etal-2024-translatotron} & 175M & 52.6 & 83.1 & 34.4 & 77.1 \\
UMTIT \cite{niu-etal-2024-umtit}& 293M & 52.0 & 80.8 & 34.2 & 78.6 \\
E2ETIT  \cite{Ma2022ImprovingET} & 122M & 31.5 & 56.1 & 19.6 & 55.8 \\
DIMTDA \cite{liang-etal-2024-document} & 242.6M & 46.6 & 82.4 & 30.5 & 71.8 \\
AnyTrans \cite{qian-etal-2024-anytrans} & - & 63.8 & 83.9 & 35.7 & 80.5 \\
\bottomrule
\end{tabular}
    }
\end{table}
%%%%%%%%%%%%%%%%%%%%%%%%%%%%%%%%%%

To validate the effectiveness of \methodname, we conducted comparative experiments against several SOTA E2E web image translation models in ZH-EN and EN-IT tasks. 
We also list the trainable parameters (\#param) for a fair efficiency evaluation.

As shown in Table~\ref{tab:translation_sota_results}, \methodname~significantly outperforms all existing methods in both tasks. 
In the ZH-EN task, \methodname~achieves 65.9 BLEU and 94.8 COMET, surpassing the previous best model, Translatotron-V, by 13.3 BLEU points and 11.7 COMET, respectively. 
In the EN-IT task, \methodname~also achieves a leading performance with 38.1 BLEU and 85.9 COMET.
The results demonstrate our method's exceptional ability to comprehend and translate complex Web images, such as those from Chinese e-Commerce sites, and to better handle the intricate interplay between vision and text in real-world Web scenarios.
These significant improvements are attributable to the two core design principles of our framework. 
First, by leveraging a powerful pre-trained LLM as a multilingual knowledge base, our framework capitalizes on its extensive linguistic priors, which is more effective than training from scratch or relying on limited task-specific data. 
Most critically, the DSAM module enables a deep synergy between semantics and details, while the VAA module dynamically injects this rich visual understanding into the LLM.
Furthermore, the comparison of parameter efficiency highlights the advantages of \methodname, our model achieves its superior performance with only approximately 50M trainable parameters, surpassing models with much larger parameter counts, such as Translatotron-V (175M) and UMTIT (293M). 
This comparison provides strong evidence that \methodname~maintains SOTA performance while also demonstrating excellent parameter efficiency, thus validating the advanced and efficient of our design.

\subsection{Ablation Study}

%%%%%%%%%%%%%%%%%%%%%%%%%%%%%%%%%%%%%%%%%%
\begin{table}[t]
\centering
\caption{Ablation Study of core components (LLM Backbone: \qwen-8B) in ZH-EN and EN-IT tasks.}
\label{tab:ablation_study_detailed}

\resizebox{0.85\linewidth}{!}{
\begin{tabular}{lcccc}
\toprule
& \multicolumn{2}{c}{\textbf{ZH-EN}} & \multicolumn{2}{c}{\textbf{EN-IT}} \\
\cmidrule(lr){2-3} \cmidrule(lr){4-5}
 & BLEU & COMET & BLEU & COMET \\
\midrule
\textbf{\methodname~(Ours)} & \bf 65.9 & \bf  94.8 & \bf 38.1 & \bf 85.9 \\
\hline
w/o DSAM  &  63.7 & 89.4 & 33.5 & 80.0 \\
w/o VAA  & 65.0 & 89.8 & 34.4 & 80.4 \\
w/o Both  & 61.3 & 88.0 & 31.9 & 78.3 \\
\bottomrule
\end{tabular}%
}
\end{table}
%%%%%%%%%%%%%%%%%%%%%%%%%%%%%%%%%%%%%%%%%%

%%%%%%%%%%%%%%%%%%%%%%%%%%%%%%%%%%%%%%%%%%%%%%%%
\begin{figure}[t]
    \centering
    \begin{subfigure}[b]{0.49\columnwidth}
        \includegraphics[width=\textwidth]{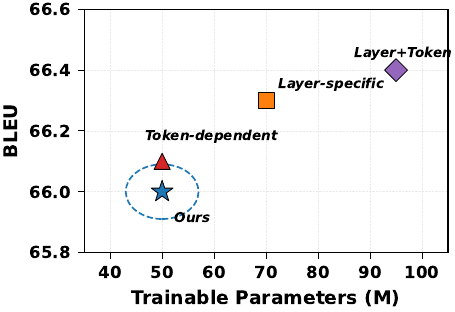}
        \caption{}
    \end{subfigure}
    % \hfill
    \begin{subfigure}[b]{0.49\columnwidth}
        \includegraphics[width=\textwidth]{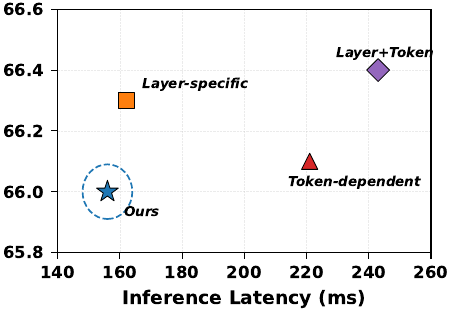}
        \caption{}
    \end{subfigure}
    \caption{Efficiency-performance trade-off of gating strategies analysis.}
    \label{fig:gating_tradeoff}
\end{figure}
%%%%%%%%%%%%%%%%%%%%%%%%%%%%%%%%%%%%%%%%%%%%%%%%

% \subsubsection{The Impact of Component.}
To investigate the contribution of each component within our framework, we conducted an ablation study focusing on DSAM and VAA.
Experiments were performed on the ZH-EN and EN-IT tasks.
We evaluated three ablated variants of \methodname:
(I) w/o DSAM: The dual-stream attention fusion is replaced by a simple concatenation of semantic and fine-grained visual features followed by an MLP.
(II) w/o VAA: The DSAM module is retained, but the Visual-Aware Adapters within the LLM are removed, disabling dynamic vision-aware adaptation.
(III) w/o Both: Both DSAM and VAA are removed, which approximates a standard LVLM built on a frozen LLM backbone.

As shown in Table~\ref{tab:ablation_study_detailed}, removing either component consistently reduces performance in both datasets.
Specifically, omitting DSAM leads to a drop of 2.2 BLEU and 5.4 COMET in ZH-EN, indicating that naive feature fusion cannot capture the complementary relationship between semantic context and orthographic details.
This result confirms that the joint modeling of textual semantics and fine-grained visual cues is crucial for accurate recognition and translation of embedded text.
Similarly, removing VAA results in a notable decrease (–0.9 BLEU and –5.0 COMET in ZH-EN), as the LLM backbone loses its ability to dynamically regulate visual influence.
Without VAA, the model struggles with visually ambiguous or noisy Web images.
The greatest degradation occurs when both modules are removed (-4.6 BLEU and -6.8 COMET in ZH-EN), demonstrating that the two components work synergistically. 
DSAM enhances the informativeness of visual representations, while VAA ensures their effective integration within the LLM.

Overall, these findings validate our design motivation: the integration of dual visual encoders and dynamic adaptation mechanisms effectively bridges the modality gap between fine-grained visual features form and multilingual semantics, leading to robust Web image translation performance.

%%%%%%%%%%%%%%%%%%%%%%%%%%%%%%%%%%%%%%%%
\begin{table}[t]
\centering
\caption{Results of different strategies for fusing visual features from \msiglip~and \dino~(LLM Backbone:\qwen-8B) in ZH-EN and HI-EN tasks.}
\label{tab:fusion_strategy_comparison}
\resizebox{0.9\linewidth}{!}{
\begin{tabular}{lcccc}
\toprule
 & \multicolumn{2}{c}{\textbf{ZH-EN}} & \multicolumn{2}{c}{\textbf{HI-EN}} \\
\cmidrule(lr){2-3} \cmidrule(lr){4-5} % Lines under ZH-EN and DE-EN
  & BLEU & COMET & BLEU & COMET \\
\midrule
\textbf{DSAM (Ours)}      & \bf 65.9 & \bf  94.8 & \bf  41.3 & \bf  85.0 \\
\midrule % Separate Ours from baselines
Simple Concat  & 63.7 & 89.4 & 38.5 & 84.8 \\
Element-wise Sum& 63.3 & 90.8 & 37.8 & 84.1 \\
Interleaving Fusion  &  63.8 & 90.9 & 38.1 & 84.4 \\
One-way CA (Sem$\to$Det) & 64.5 & 91.0 & 39.2 & 84.9 \\
One-way CA (Det$\to$Sem) & 64.2 & 90.7 & 38.9 & 84.8 \\
Self-Attention    & 65.0 & 91.8 & 39.7 & 84.9 \\
\bottomrule
\end{tabular}%
} 
\end{table}
%%%%%%%%%%%%%%%%%%%%%%%%%%%%%%%%%%%%%%%%

\section{Analysis}

\subsection{Gating Strategy Analysis}
To validate the VAA gating design, we compare different gating strategies, evaluating their trade-offs in performance, parameter efficiency, and inference speed.
We implement four gating strategies on the IT-EN task:
(I) Global Gating (Ours): All layers share a single gating vector.
(II) Layer-specific Gating: Each layer independently computes its gating vector, allowing different layers to have varying degrees of visual dependence.
(III) Token-dependent Gating: Gate values are dynamically computed based on each token position's LLM hidden state, enabling the model to apply different visual weights to different tokens.
(IV) Layer+Token Gating: Combines both strategies.
Figure~\ref{fig:gating_tradeoff} illustrates the trade-off between performance and resource consumption. The left panel shows the relationship between BLEU scores and trainable parameters, while the right panel shows BLEU scores versus inference latency.
Layer-specific gating provides minimal performance improvement while requiring 40\% more parameters. Token-dependent gating yields almost no performance gain while substantially increasing inference latency by 42\%.
The most complex variant, which combines both layer and token dimensions, achieves the highest performance, but at the cost of 90\% more parameters and 56\% higher latency. 
The performance improvement is disproportionate to the resource consumption (nearly doubled). Such extreme trade-offs are impractical for real-world deployment scenarios.
Our gating design maintains competitive performance, only 0.4 BLEU below the best configuration, while consuming nearly half the resources. 
These results validate our gating optimality in both the performance parameter and performance-latency dimensions.

\subsection{Vision Feature Fusion Strategies}
\label{sec:fusion_comparison}

The effective fusion of complementary features from the \msiglip~and \dino~ encoders is crucial in our visual encoder. 
In this section, we compare DSAM in \methodname~framework with several baseline on ZH-EN and HI-EN tasks.
We evaluated the following fusion strategies. 
(I) Simple Concat: Features are concatenated along the feature dimension ($\text{Concat}(\mathbf{H}_\text{s}, \mathbf{H}_\text{d})$) and then processed by a 2-layer MLP. 
(II) Element-wise Sum: Features are added element-wise ($\mathbf{H}_\text{s} + \mathbf{H}_\text{d}$) and then processed by a 2-layer MLP.
(III) Interleaving Fusion: Features are interleaved along the sequence dimension, creating a sequence of length $2\mathbf{N}_\text{d}$, which is then processed by a 2-layer MLP.
(IV) One-way Cross-Attention (Sem$\to$Det): Semantic features serve as queries to attend to detail features via cross-attention, followed by LayerNorm and an MLP. The hidden dimensions are adjusted so that the total parameter count of the fusion module is comparable to DSAM.
(V) One-way Cross-Attention (Det$\to$Sem): The reverse direction, where detail features query semantic features.
(VI) Self-Attention: Features are concatenated ($\text{Concat}(\mathbf{H}_\text{s}, \mathbf{H}_\text{d}))$) and then fed into a standard Transformer Encoder layer to allow interaction via self-attention, followed by an MLP for dimension adjustment.

As shown in Table~\ref{tab:fusion_strategy_comparison}, we compare the performance of different strategies to fuse the visual features of \msiglip~and \dino. 
Compared to Simple Concat, DSAM yields improvements of nearly 3 BLEU on both ZH-EN and HI-EN tasks, demonstrating that the basic feature combination fails to fully exploit the complementary information. 
Although more sophisticated methods like Interleaving Fusion and Self-Attention show gains over the simplest baselines, they still fall considerably short of DSAM.

To further isolate the necessity of bidirectional interaction, we introduce two parameter-matched one-way cross-attention variants, where the fusion module parameters are controlled to be comparable to DSAM by adjusting hidden dimensions. 
As shown in Table~\ref{tab:fusion_strategy_comparison}, One-way CA (Sem$\to$Det) achieves 64.5/91.0 and One-way CA (Det$\to$Sem) achieves 64.2/90.7 (BLEU/COMET) on ZH-EN, both trailing DSAM by 1.4--1.7 BLEU and 3.8--4.1 COMET. 
A similar pattern is observed on HI-EN, where both one-way variants (39.2 and 38.9 BLEU) lag behind DSAM (41.3 BLEU) by over 2 BLEU points. 
Notably, Self-Attention, which allows implicit bidirectional interaction through self-attention over concatenated features, also underperforms DSAM, suggesting that the explicit and structured bidirectional cross-attention in DSAM where features from one encoder directly query and attend to features from the other is more effective than applying self-attention to already mixed features.

We attribute DSAM's superiority to its explicit bidirectional enhancement mechanism: semantic context guides the refinement of fine-grained visual features (Sem$\to$Det), while detailed visual cues simultaneously enrich semantic clarity (Det$\to$Sem). 
This mutual refinement produces a unified representation that is strictly more informative than what either direction alone can achieve, which is critical to accurately recognizing and translating the diverse text styles found in Web images.

\subsection{VAA Insertion Strategies} 
\label{sec:adapter_insertion_final}

%%%%%%%%%%%%%%%%%%%%%%%%%%%%%%%%%%%%%%%%

\begin{table}[t]
\centering
\caption{Results of different VAA insertion strategies in JA-EN and HI-EN tasks (LLM Backbone: \qwen-8B).}
\label{tab:adapter_insertion_comparison_final}
\resizebox{0.95\columnwidth}{!}{% % Uncomment if table is too wide
\begin{tabular}{lcccc}
\toprule
 & \multicolumn{2}{c}{\textbf{JA-EN}} & \multicolumn{2}{c}{\textbf{HI-EN}} \\
\cmidrule(lr){2-3} \cmidrule(lr){4-5} 
& BLEU & COMET & BLEU & COMET \\
\midrule
\textbf{Uniform Insertion(Ours)} & \textbf{47.2}   & \textbf{86.6}    & \textbf{41.3}   & \textbf{85.0}    \\
\midrule
Early Insertion   &   45.1            &   84.9             &   39.0            &   83.1             \\
Late Insertion &   46.5            &   85.8             &   40.5            &   84.2             \\
\bottomrule
\end{tabular}%
} % Uncomment if using resizebox
\end{table}

The VAA is a key component to enhance the LLM's adaptability for the web image translation task. 
To investigate the optimal insertion strategy, we compared the performance differences arising from inserting adapters at various locations within the LLM backbone.
We evaluated three primary strategies.
\textbf{Uniform Insertion (Ours)} Adapters are inserted after the FFN sub-layer in all layers. 
\textbf{Early Insertion} Adapters are inserted only into the first 12 layers.
\textbf{Late Insertion} Adapters are inserted only into the last 12 layers.
To ensure a fair comparison despite the varying number of adapters, we adjusted the bottleneck dimension for the Early and Late Layers strategies so that their total number of trainable adapter parameters closely matched that of the Uniform Insertion strategy. 
We evaluated these strategies in the JA-EN and HI-EN tasks.

As shown in Table~\ref{tab:adapter_insertion_comparison_final},  we compare the performance of these different adapter insertion strategies with controlled parameter counts. 
In contrast, inserting adapters only into the early or late stages leads to performance degradation. 
Specifically, the Early Layers strategy exhibited the lowest performance, with BLEU dropping by 2.1 and 2.3 points on JA-EN and HI-EN, respectively, compared to Uniform Insertion. 
While the Late Layers strategy performed better than Early Layers, it still lagged significantly behind the uniform approach.
These findings suggest that effective adaptation for web image translation requires adjustments throughout the LLM's entire processing depth, even when parameter budgets are matched. 
Modifying only early layers appears insufficient to refining complex semantic representations and generation decisions made in later stages. 
In contrast, adapting only late layers misses the opportunity to integrate visual guidance during earlier feature processing and representation learning phases.

\subsection{The Vision Encoder}

%%%%%%%%%%%%%%%%%%%%%%%%%%%%%%%%%%%%%%%%%%%%%%%%
\begin{table}[t]
\centering
\caption{Results on different combinations of Vision Encoder (LLM Backbone: \qwen-8B) in ZH-EN and HI-EN tasks.}
\label{tab:encoder_ablation_detailed}
\resizebox{0.85\linewidth}{!}{
\begin{tabular}{lcccc}
\toprule
\multirow{2}{*}{\textbf{Vision Encoder}} & \multicolumn{2}{c}{\textbf{ZH-EN}} & \multicolumn{2}{c}{\textbf{HI-EN}} \\
\cmidrule(lr){2-3} \cmidrule(lr){4-5} % Lines under ZH-EN and HI-EN
& BLEU & COMET & BLEU & COMET \\
\midrule
\textbf{\msiglip~+ \dino~(Ours)}         & \bf 65.9 & \bf  94.8 & \bf  41.3 & \bf  85.0 \\
\midrule
Only \msiglip~ & 61.4 & 90.5 & 40.6 & 84.3 \\
Only \dino~  & 60.3          & 89.1           & 31.8          & 81.5           \\
\midrule % Separate single from alternative dual
\clip~ + \dino~   & 60.5 & 89.3 & 38.5 & 81.6 \\
\msiglip~+ MAE    & 63.8 & 90.8 & 40.9 & 84.2 \\
\bottomrule
\end{tabular}%
} 
\vspace{-2mm}
\end{table}

%%%%%%%%%%%%%%%%%%%%%%%%%%%%%%%%%%%%%%%%%%%%%%%%

To further examine the effectiveness and generalizability of our DSAM design, we conducted an ablation study by varying the combination of visual encoders while keeping the rest of the architecture fixed.
Beyond the baselines with the single vision encoder, we evaluated the performance when replacing \msiglip~or \dino~ with other representative encoders such as \clip~(ViT-L/14) \footnote{\url{https://huggingface.co/openai/clip-vit-large-patch14}} and MAE (ViT-L/16)\footnote{\url{https://huggingface.co/facebook/vit-mae-large}}. 
All variants of visual encoders employed DSAM.

As shown in Table~\ref{tab:encoder_ablation_detailed}, we have three key findings.
The Visual Encoders consistently outperform single-encoder baselines (Only \msiglip~or Only \dino), confirming that integrating complementary visual information is essential for accurate Web image translation. 
Both are indispensable for recognizing stylized or noisy text in real-world Web scenes.
Among dual-stream combinations, \msiglip~and \dino~achieve the best overall performance, surpassing all alternatives by a clear margin (e.g., +2.1 BLEU / +4.0 COMET over \msiglip~and MAE in ZH-EN).
This indicates that the synergy of visual features between the two encoders is especially effective in bridging the modality gap between fine-grained visual features and multilingual semantics.
Importantly, other combinations such as \clip~and \dino~and \msiglip~and MAE also show notable gains compared to single-encoder setups, validating the generalizability of our framework.
Our DSAM does not depend on a specific encoder pair, it can flexibly integrate diverse visual encodes, maintaining performance stability across architectures.
This adaptability demonstrates that \methodname~is not limited to a particular model choice, but can serve as a general plug-and-play framework for web image translation tasks.

\subsection{Sensitivity Analysis of  \texorpdfstring{$\lambda$}{lambda} in Stage 2}
%%%%%%%%%%%%%%%%%%%%%%%%
\begin{table}[t]
\centering
\caption{Results of different combinations of loss weight in Multi-task Joint Learning Stage (Stage 2). We evaluate performance on ZH-EN task (LLM Backbone: \qwen-8B).}
\label{tab:loss_weights_ablation}
\resizebox{0.75\linewidth}{!}{
\begin{tabular}{lcc}
\toprule
\small \textbf{Loss Weight (ITM, TTL, ITL)} & \small \textbf{BLEU} & \small \textbf{COMET} \\
\midrule
\textbf{\methodname~(Ours)} & \textbf{65.9} & \textbf{94.8}\\
\midrule
(0.2, 0.3, 0.5) & 65.5 & 94.5\\
(0.3, 0.3, 0.4) & 65.7 & 94.6 \\
(0.4, 0.2, 0.4) & 65.6 & 94.5 \\
\bottomrule
\end{tabular}
}
\end{table}

In this section, we conducted a sensitivity analysis on the loss weights of Multi-Task Joint Learning Stage (Stage 2).
We compared our adopted weight combination (0.3, 0.2, 0.5) with several other plausible configurations on ZH-EN task.
The results are presented in Table \ref{tab:loss_weights_ablation}.
As can be seen in the table, although different weight combinations show minor differences in BLEU and COMET, overall performance is relatively robust after the model is fully trained, without significant fluctuations. 
This indicates that our \methodname~framework is not extremely sensitive to minor adjustments in loss weights.
Overall, our currently selected weights demonstrated stable and comprehensive performance in our experiments, confirming the soundness and reliability of our methodology.

\subsection{Training Schedule Sensitivity Analysis}

VaaWIT adopts a two-stage training paradigm: Stage~1 for visual-language alignment and Stage~2 for multi-task joint learning, each trained for one epoch. To investigate the sensitivity of this design, we conduct ablations on the ZH-EN task by varying the number of training epochs per stage and by skipping Stage~1 entirely.

\begin{table}[t]
\caption{Sensitivity analysis of the training schedule on the ZH-EN task (LLM Backbone: Qwen3-8B). ``S1'' and ``S2'' denote Stage~1 and Stage~2, respectively, and ``ep'' denotes epochs.}
\label{tab:stage_sensitivity}
\centering
\begin{tabular}{lcc}
\toprule
Configuration & BLEU & COMET \\
\midrule
\textbf{S1 (1ep) + S2 (1ep) (Ours)} & \textbf{65.9} & \textbf{94.8} \\
\midrule
S1 (2ep) + S2 (1ep) & 66.2 & 95.0 \\
S1 (1ep) + S2 (2ep) & 66.4 & 95.1 \\
\midrule
Skip S1, S2 only (1ep) & 63.1 & 91.4 \\
Skip S1, S2 only (2ep) & 64.0 & 92.6 \\
\bottomrule
\end{tabular}
\vspace{-3mm}
\end{table}

As shown in Table~\ref{tab:stage_sensitivity}, three key findings emerge.
First, Stage~1 is indispensable. Skipping Stage~1 and directly performing Stage~2 leads to a substantial degradation of 2.8 BLEU and 3.4 COMET (65.9$\to$63.1 / 94.8$\to$91.4). 
Even doubling the Stage~2 epochs without Stage~1 (64.0 BLEU) cannot recover the performance of the full two-stage pipeline with a single epoch each (65.9 BLEU). 
This confirms that the visual-language alignment in Stage~1 provides a critical initialization for the DSAM and VAA modules that cannot be substituted by additional task-specific training alone.
Second, additional epochs yield diminishing returns. 
Extending either stage to two epochs provides only marginal gains (+0.3--0.5 BLEU, +0.2--0.3 COMET), indicating that the model converges efficiently within a single epoch per stage.
Third, these results collectively validate our 1+1 epoch schedule as an effective performance--efficiency trade-off, achieving near-optimal performance while requiring only approximately 18 hours of training on 8$\times$NVIDIA H20 GPUs.

\subsection{Case Study}
%%%%%%%%%%%%%%%%%%
\begin{figure}[t]
    \centering
    \includegraphics[width=0.95\linewidth]{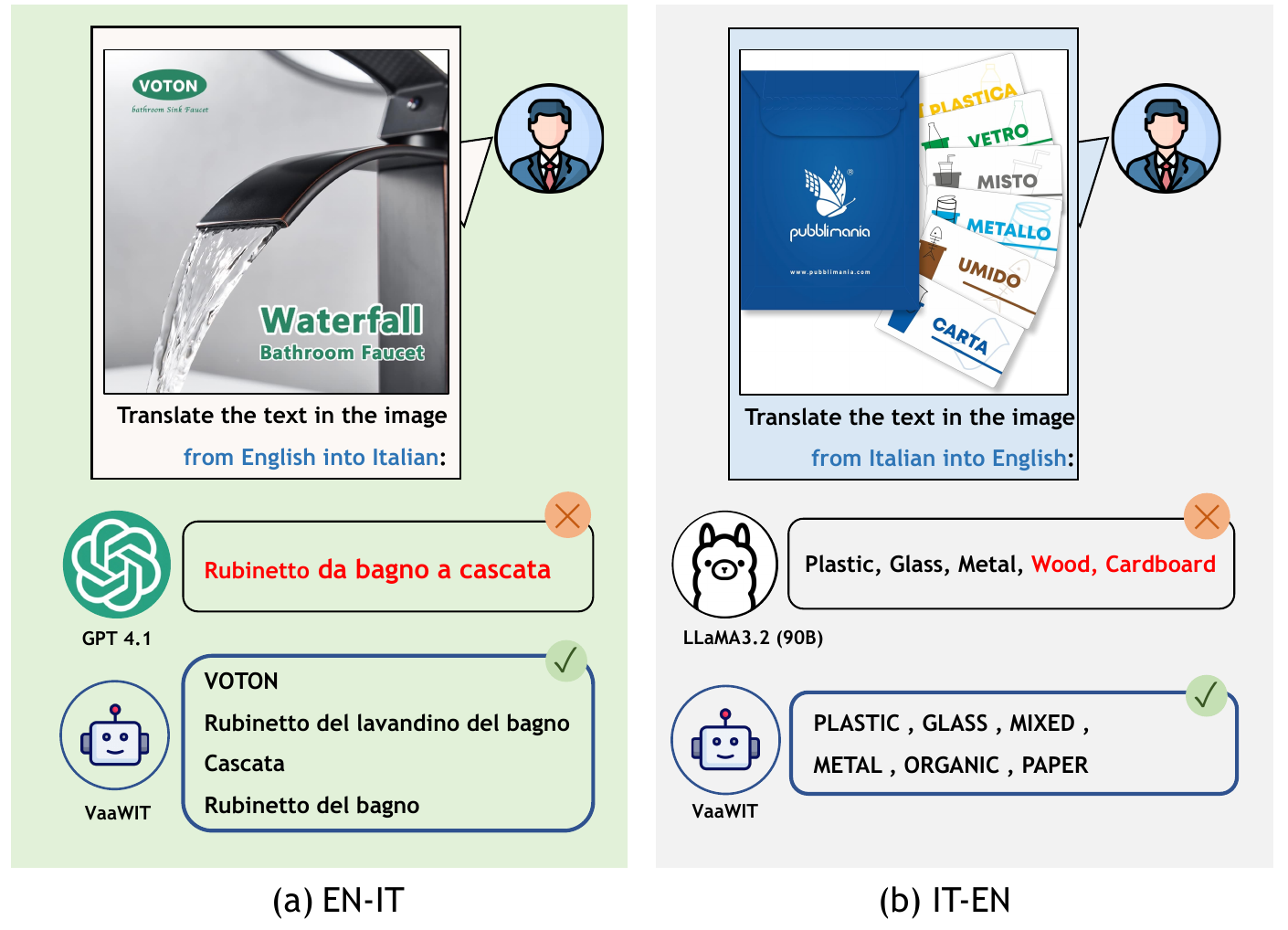}
    \caption{Case Study of \methodname~Framework.}
    \label{fig:case_study}
    % \vspace{-2mm}
\end{figure}
%%%%%%%%%%%%%%%%%%

To further validate the effectiveness of \methodname~in complex Web image translation scenarios, we present two representative cases as shown in Figure ~\ref{fig:case_study}.

\noindent \textbf{Case 1 (EN-IT):}
This case features a typical e-commerce product image, where text is scattered in different locations, mixing a brand logo with descriptive text. 
\gptfour~generated a semantically fluent translation, ``Rubinetto da bagno a cascata'', but completely omitted the brand name ``VOTON'' and parts of the descriptive phrases. 
In contrast, \methodname~provided a complete translation containing all the textual information, maintaining a high consistency with the source in both structure and semantics.
This difference illustrates a key distinction between general-purpose LVLMs and our specialized framework. 
\gptfour~is biased towards understanding the overall gist of an image. 
\methodname~explicitly integrates semantic and fine-grained visual features through deep cross-stream interaction, effectively preventing critical information loss.

\noindent \textbf{Case 2 (IT-EN):}
This image contains 6 labels. \mllama~(90B) exhibited two typical errors: information omission (failing to recognize ``MISTO'') and contextual mistranslation (misinterpreting ``UMIDO''  as ``Wood''). 
In contrast, \methodname~not only translated all labels, but also accurately leveraged the visual context to translate ``UMIDO'' as ``Organic''.
The errors made by \mllama~(90B) reveal the limitations of deep- and fine-grained cross-modal reasoning.

Overall, \methodname~demonstrates clear advantages in visually complex and text-dense Web images, achieves more accurate and robust multilingual web image translation.

%%%%%%%%%%%%%%%%%%%%%%%%%%%%%%%%%%%%%%%%%%%%%%%%
\section{Related work}
\label{sec:related_works}
 
Web image translation aims to understand and translate text embedded within Web visual content, a cross-modal task fundamentally different from text-only Neural Machine Translation \cite{sutskever2014sequence, bahdanau2014neural, cho-etal-2014-properties}.
These images often embed multilingual text in advertising, product displays, information dissemination, and user-generated content—such as e-commerce product images and social media posts—where text exhibits high diversity in fonts and colors, and contains complex multi-line structures, presenting significant barriers to information access for global users \cite{mansimov2020towards, lan-etal-2024-translatotron}.

Early Web image translation methods predominantly used cascaded systems, combining Optical Character Recognition (OCR) with Machine Translation (MT) \cite{gu2017non}.
As mentioned in the introduction, these traditional pipelines are particularly fragile when processing real-world Web images, as OCR errors lead to erroneous translation results (i.e., error propagation) \cite{yin2023multi}.
To address the limitations of cascaded methods, end-to-end (E2E) image translation models integrate visual text recognition and translation into unified architectures.
Existing E2E models have explored various optimization strategies: multi-task learning \cite{Ma2022ImprovingET}, knowledge distillation \cite{ma2023multi}, modality alignment mechanisms \cite{zhu-etal-2023-peit}, and multimodal representation learning \cite{lan2023exploring, lan-etal-2024-translatotron}.
These models provide more concise and unified approaches, but are often limited by the scale and diversity of task-specific training data, struggling to handle the complexity of multilingual and multi-domain scenarios on the Web \cite{zhu-etal-2023-peit, liang-etal-2024-document, niu-etal-2024-umtit}.
Recently, LVLMs \cite{liu2024llavanext, lu2024deepseek, chen2024internvl, llama3modelcard, li2023blip}, built upon LLMs and pre-trained on massive image-text corpora, possess powerful cross-modal understanding capabilities, offering new possibilities for Web image translation.
However, directly applying standard LVLM to Web image translation exposes two core challenges:
(1) Visual Representation Gap—mainstream LVLM visual encoders (e.g. CLIP \cite{radford2021learning}) optimize for image-level semantic understanding through contrastive learning, while Web image translation requires fine-grained visual features \cite{zhu-etal-2023-peit,li-etal-2025-mit};
(2) The Fusion and Adaptation Challenge—even with multi-source visual features, how to effectively integrate this information and enable LLMs to robustly handle the diversity of Web images remains an unresolved challenge \cite{ebrahimi2024crome}.
Recent work has explored fusing multiple visual encoders \cite{lin2023sphinx, jiang2023clip, shi2024eagle, luo2024feast}, but mainstream fusion methods (such as simple feature concatenation or gating) are too shallow, failing to achieve deep synergy between macro-level multilingual semantics and fine-grained visual details.

\methodname~addresses the unique challenges of Web image translation through the DSAM and VAA modules.
DSAM enables deep interaction between semantic and visual detail features through bidirectional cross-attention to bridge the visual representation gap. 
VAA achieves parameter-efficient and context-aware LLM adaptation through visual-aware dynamic gating.

%%%%%%%%%%%%%%%%%%%%%%%%%%%%%%%%%%%%%%%%%%%%%%%%%%%%%%%%%%%%%%%%%%%%%%%%%%%%%%%
\section{Conclusion}
\label{sec:conclusion}

In this paper, we present \methodname, a novel framework specifically designed for the challenging task of multilingual web image translation.
The strength of \methodname lies in the synergy of two key innovations: the DSAM, which enables deep interaction between complementary visual features, and the VAA, which facilitates parameter-efficient, dynamic adaptation within the LLM backbone. 
Through extensive experiments on several public benchmarks, our approach has not only outperformed previous methods but also achieved performance comparable to commercial systems. 
Our work demonstrates the powerful potential of combining structured visual feature fusion with dynamic, lightweight LLM adaptation for complex cross-modal tasks, while maintaining remarkable parameter and training efficiency. 
Future work could explore extending \methodname with layout-aware translation.

\section{Acknowledgements}
This work was supported by the National Key Research and Development Program of China (Grant No. 2024YFB3309702) and the National Natural Science Foundation of China Youth Foud (Grant No. 62306210).

\clearpage
\bibliographystyle{ACM-Reference-Format}
\bibliography{refs}

\appendix

%%%%%%%%%%%%
\begin{table}[t]
\centering
\caption{Dataset Statistics.}
\resizebox{0.85\linewidth}{!}{
\begin{tabular}{l|lll}
\toprule
\textbf{Dataset} & \textbf{Train} & \textbf{Test} & \textbf{Tasks} \\ \midrule
\datasetecoit & 480K  & 100 & ZH-EN          \\
% \datasetiitm  & 210K  & 200 & DE-EN, FR-EN   \\
\datasetmit   & 800K  & 400 & EN-IT, EN-JA, IT-EN, JA-EN   \\
\datasetopus  & 876K  & 300 & HI-EN, KO-EN, TH-EN \\ 
\bottomrule
\end{tabular}
}

\label{tab:dataset_statistics}
\end{table}
%%%%%%%%%%%%
% ################
\begin{table}[t]
\centering
\caption{Training Hyperparameters Configuration.}
\begin{tabular}{l|ccc}
\toprule
\textbf{Hyperparameters} & \textbf{Stage 1}  & \textbf{Stage 2} \\
\midrule
Learning rate & 1.00E-03  & 1.00E-05 \\
LR scheduler & Cosine  & Cosine \\
Weight decay & 3.00E-06  & 3.00E-06 \\
Gradient clip & 1.0  & 1.0 \\
Optimizer & \multicolumn{3}{c}{AdamW($\beta_1 = 0.8$, $\beta_2 = 0.95$)} \\
Warm-up & 0.08  & 0.03 \\
Batch size & 256  & 512 \\
Sequence length & 2048 &  2048 \\
Epochs & 1 & 1  \\
\bottomrule
\end{tabular}
\label{tab:hyperparameters}
\end{table}
% ################

\section{Implementation Details}
\label{sec:implementation_details}

The operating system which we use is CentOS release 7.5, and the programming language is Python 3.9.12. 
Our experiments were conducted on NVIDIA H20 GPUs, the CUDA version is 12.2, and the deep learning framework is torch with version 2.3.1, torchvision with version 0.18.1 and Transformers with 4.57.0.
%”All experiments for our proposed \methodname~framework utilize the \textbf{
% \qwen-3B}\footnote{\url{https://huggingface.co/Qwen/Qwen2.5-3B}} as the frozen LLM backbone. 
For the visual encoder, we employ \textbf{\msiglip} and \textbf{\dino}. 
Both visual encoders are also kept frozen during training. 

\subsection{Datasets}
We conducted experiments on three datasets of public Web image translation, with statistics shown in Table~\ref{tab:dataset_statistics}.

\subsection{Two-Stage Training Settings}

\textbf{Stage 1: Visual-Language Alignment.}
In this stage, we use the complete training data from all three datasets (\datasetmit, \datasetecoit, \datasetopus) for visual-language alignment training. 
Specifically, the input data consist of image content and the source language text in the image. 
Through auto-regressive language modeling loss (detailed in Section~\ref{sec:training}), we train the DSAM and VAA modules to align fused visual representations with textual semantics in the LLM embedding space.

\noindent \textbf{Stage 2: Multi-Task Joint Learning.}
In this stage, we construct a mixed training set containing data for three complementary tasks:
First, we randomly sample 30\% of the Stage 1 data to continue alignment training and maintain visual-language correspondence.
Second, we used source texts and target translations from the three dataset training sets  to train pure text translation capability.
Then, we used the complete data from the three training datasets (image, source text, and target translation) to train end-to-end image translation capability.
Data from the three tasks are mixed-sampled in each training batch according to loss weights $\lambda_{\text{ITM}} = 0.3$, $\lambda_{\text{TTL}} = 0.2$, $\lambda_{\text{ITL}} = 0.5$ (detailed in \S \ref{sec:training}) to form the final training data stream.

Table~\ref{tab:hyperparameters} lists the hyperparameter configurations for both training stages.
All experiments were conducted on a single node with 8 NVIDIA H20 GPUs (96GB HBM3 each). 
We use DeepSpeed ZeRO Stage 2 for data parallelism with gradient accumulation. 
Only the DSAM and VAA modules (approximately 50M parameters) are updated during training, while the visual encoders and LLM backbone remain frozen. Stage 1 training requires approximately 6 hours (1 epoch), and Stage 2 training requires approximately 12 hours (1 epoch), totaling approximately 18 hours.

\subsection{Baselines and Fairness of Comparison}
To ensure fair and reproducible comparisons, all methods are evaluated using the same test sets, metrics, and decoding settings. The following protocols are strictly followed:

\noindent \textbf{Zero-shot LVLM Baselines.}
All zero-shot LVLMs are evaluated with the unified prompt: ``\textit{Translate the text in the image from [Source Language] into [Target Language]:}'', using original-resolution images, greedy decoding (temperature$=$0, max 512 tokens), with OCR/tool-use options disabled. Commercial model evaluations (GPT-4.1, Gemini~2.5~Pro) were conducted in October 2025.

\noindent \textbf{Tuning Strategy Baselines.}
LoRA ($r{=}8$, 30M parameters) and Full Fine-Tuning (8B parameters) are applied to Qwen3-VL (8B) using the exact same Stage~2 training data as \methodname. CoT uses Qwen3-VL (8B) with prompting only.

\noindent \textbf{SOTA E2E Image Translation Models.}
All E2E baselines are reproduced using official code, retrained on our combined training dataset with original hyperparameters: ItNet (60.6M), PEIT (ResNet variant, 71.6M), E2ETIT (122M), Translatotron-V (175M), DIMTDA (242.6M), and UMTIT (293M). 
AnyTrans is training-free and evaluated using its published pipeline.

\section{Further Analysis}
%%%%%%%%%%%%%%%%%%%%%%%%
\begin{table}[t]
\centering
\small
\caption{Performance degradation under different noise conditions on ZH-EN task. All degradation percentages are relative to baseline.}
\label{tab:noise_robustness}
\begin{tabular}{ll|l}
\toprule
\textbf{Noise Type} & \textbf{Ours} & \textbf{Full FT} \\
\midrule
Baseline & 65.9 & 61.8 \\
\midrule
Gaussian Blur ($\sigma=2$) & 62.3 (-5.5\%) & 56.4 (-8.7\%) \\
JPEG Compression (Q=30) & 63.8 (-3.2\%) & 58.1 (-6.0\%) \\
Low Resolution (50\%) & 61.7 (-6.4\%) & 55.2 (-10.7\%) \\
Occlusion (15\% area) & 63.1 (-4.2\%) & 57.9 (-6.3\%) \\
% \midrule
Mixed Noise & 59.2 (-10.2\%) & 52.0 (-15.8\%) \\
\bottomrule
\end{tabular}
\end{table}

\subsection{Robustness Analysis}
\label{sec:noise_robustness} 

Real-world web images frequently suffer from various degradation factors, including compression artifacts from social media platforms, low-resolution captures from mobile devices, motion blur, and partial occlusions from watermarks or overlays. 
To validate \methodname's robustness in practical deployment scenarios, we conducted systematic noise robustness evaluation on the ZH-EN task using the \datasetmit~test set.

We simulated five common types of web image degradation to comprehensively assess model robustness:
(I) Gaussian Blur: Simulates camera defocus or motion blur during capture.
(II) JPEG Compression (quality = 30): Simulates aggressive compression used by social media platforms to reduce bandwidth.
(III) Low Resolution (50\% downsampling): Simulates images captured by low-end devices or generated as thumbnails.
(IV) Occlusion (15\% random area): Simulates watermarks, user interface overlays, or partial content damage.
(V) Mixed Noise: Applies all the above degradations simultaneously, representing the most realistic and challenging web scenario.
For each noise type, we applied the degradation to all test images and evaluated both \methodname~and the Full Fine-Tuning baseline to measure the relative performance degradation.

As shown in Table~\ref{tab:noise_robustness}, \methodname~shows superior robustness compared to the baseline for all individual noise types. The performance gap is especially pronounced under low-resolution conditions, where the' degradation of \methodname (-6.4\%) is significantly lower than that of Full FT (-10.7\%).
Under the most challenging mixed noise conditions, which most closely simulate real-world web scenarios where multiple degradation factors co-occur, \methodname~maintains 59.2 BLEU with only 10.2\% degradation, while Full FT drops to 52.0 BLEU (-15.8\%).
This 5.6 percentage point difference in robustness demonstrates the practical value of \methodname's design for real-world deployment.

\subsection{VAA's Parameter Effects Analysis}

\begin{table}[t]
\centering
\caption{VAA's parameter effects analysis in VAA on ZH-EN task.}
\label{tab:vaa_mechanism}
\resizebox{0.9\columnwidth}{!}{
\begin{tabular}{lcc}
\toprule
 & \textbf{BLEU} & \textbf{COMET}\\
\midrule
w/o VAA (Baseline) & 65.0 & 89.8  \\
\midrule
Random Gate & 64.5 (-0.5) & 89.1(-0.7)  \\
Fixed Gate ($g=0.5$) & 64.8 (-0.2) & 89.5 (-0.3) \\
Fixed Gate ($g=1.0$) & 65.3 (+0.3) & 90.2 (+0.4) \\
\midrule
\textbf{Ours (Dynamic Gate)} & \textbf{65.9} (\textbf{+0.9})& \textbf{94.8} (\textbf{+5.0}) \\
\bottomrule
\end{tabular}
}
\end{table}

A critical question in adapter-based fine-tuning is whether performance gains arise primarily from additional trainable parameters or from the specific adaptation mechanism itself. 
To rigorously address this question for our VAA, we designed experiments that isolate the contribution of dynamic gating from the effect of added parameters.
We constructed four comparative configurations on the ZH-EN task, all sharing identical base architectures but differing in their gating strategies:
(I) w/o VAA: Completely removes the adapter module, serving as the baseline. The model relies solely on visual features fused with DSAM without any adaptation mechanism.
(II) Fixed Gate ($g=1.0$): Retains the complete adapter structure (down-projection, ReLU, up-projection) with parameters identical to \methodname, but the gating vector is fixed to $g=1.0$ (fully open). 
This isolates the effect of additional parameters without dynamic modulation.
(III) Fixed Gate ($g=0.5$): Same adapter parameters as above, but with gating fixed to $g=0.5$ (50\% weight). 
This tests whether a middle-ground static strategy can approximate dynamic behavior.
(IV) Random Gate: Adapter parameters identical to \methodname, but gating values are randomly sampled from $\mathcal{N}(0.5, 0.1)$ at each forward pass. 
This verifies whether visual awareness is necessary or if arbitrary modulation suffices.
(V) Ours (Dynamic Gate): Our complete approach with visual-aware dynamic gating, where $g = \sigma(\text{MLP}_G(h_g))$ adapts based on global visual features.

Table~\ref{tab:vaa_mechanism} shows the performance across different configurations. 
Comparing Fixed Gate ($g=1.0$) with w/o VAA reveals the contribution of adapter parameters alone: +0.3 BLEU (65.0$\rightarrow$65.3) and +0.4 COMET. 
This accounts for only 33\% of the total performance gain, demonstrating that simply adding adapter parameters provides limited benefit.
The gap between \methodname~and Fixed Gate ($g=1.0$) isolates the pure contribution of the dynamic gate mechanism: +0.6 BLEU (65.3$\rightarrow$65.9) and +4.6 COMET. 
This represents 67\% of the total gain, conclusively showing that the adaptive modulation mechanism is the primary driver of VAA's effectiveness.
The Random Gate configuration, despite having the same parameters as \methodname, performs worse than the w/o VAA baseline (64.5 vs. 65.0 BLEU). 
This counterintuitive result shows that arbitrary modulation actively harms performance. 
The gating mechanism must be visual-aware to provide benefit. Random or uninformed modulation introduces noise that disrupts the LLM's internal processing.
Similarly, Fixed Gate ($g=0.5$), though representing a reasonable middle-ground strategy, underperforms the baseline. 
This indicates that a static compromise cannot substitute for adaptive, context-dependent modulation.

\subsection{Fine-grained Text Fidelity Analysis}

While BLEU and COMET are widely used for evaluating translation quality, they may not fully capture character-level text recognition fidelity, which is critical for Web image translation involving brand names, numeric tokens, and special characters. To provide a more comprehensive evaluation, we report Character Error Rate (CER) and numeric token accuracy on the ZH-EN task, comparing VaaWIT against representative baselines.

\begin{table}[t]
\caption{Fine-grained text fidelity metrics on ZH-EN task (LLM Backbone: Qwen3-8B). CER measures character-level recognition errors (lower is better). Numeric Accuracy measures the proportion of correctly translated numeric tokens (higher is better).}
\label{tab:fidelity}
\centering
\begin{tabular}{lcc}
\toprule
 & CER $\downarrow$ & Numeric Acc. $\uparrow$ \\
\midrule
Qwen3-VL-32B (Zero-Shot) & 28.7\% & 65.3\% \\
Full Fine-Tuning & 12.3\% & 81.2\% \\
\midrule
\bf \methodname~(Ours) & \textbf{8.1\%} & \textbf{89.7\%} \\
\bottomrule
\end{tabular}
\end{table}

As shown in Table~\ref{tab:fidelity}, \methodname~achieves 8.1\% CER and 89.7\% numeric accuracy, outperforming Full Fine-Tuning by 4.2 and 8.5 percentage points, respectively. 
The improvement is particularly significant compared to zero-shot Qwen3-VL-32B, which suffers from a CER of 28.7\% and numeric accuracy of only 65.3\%. 
These results validate that the DSAM module's fine-grained visual fusion directly enhances character-level recognition, especially for the stylized text, brand logos, and numeric information commonly found in e-commerce Web images. 
The strong numeric accuracy further confirms \methodname's ability to preserve critical quantitative information (e.g., prices, quantities, specifications) during translation, which is essential for real-world Web content accessibility.

\end{document}